\documentclass[sigconf]{acmart}
\AtBeginDocument{%
  }

\setcopyright{acmcopyright}
\copyrightyear{2018}
\acmYear{2018}
\acmDOI{XXXXXXX.XXXXXXX}

\acmConference[CPSS 2023]{9th ACM Cyber-Physical System Security Workshop}{July 10,
  2022}{Melbourne, Australia}
\acmPrice{15.00}
\acmISBN{978-1-4503-XXXX-X/18/06}

\usepackage{amsmath}
\usepackage{listings}
\usepackage{xcolor}
\usepackage{tikz}
\usepackage{multirow}
\usepackage{adjustbox}
\usepackage{subfig}
\usepackage{gensymb}

\definecolor{codegreen}{rgb}{0,0.6,0}
\definecolor{codegray}{rgb}{0.5,0.5,0.5}
\definecolor{codepurple}{rgb}{0.58,0,0.82}
\definecolor{backcolour}{rgb}{0.95,0.95,0.92}

\lstdefinestyle{mystyle}{
    commentstyle=\color{codegreen},
    keywordstyle=\color{magenta},
    numberstyle=\tiny\color{codegray},
    stringstyle=\color{codepurple},
    basicstyle=\ttfamily\footnotesize,
    breakatwhitespace=false,         
    breaklines=true,                 
    captionpos=b,                    
    keepspaces=true,                 
    numbers=left,                    
    numbersep=5pt,                  
    showspaces=false,                
    showstringspaces=false,
    showtabs=false,                  
    tabsize=2
}

\tikzset{every picture/.style={line width=0.75pt}} 

\begin{document}



\lstdefinelanguage{ST}
{
	morekeywords={
	case,of,if,then,end_if,end_case,super,function_block,function,extends,var,
	constant, byte,,end_var,var_input, real,bool,var_output,
	dint,udint,word,dword,array, of,uint,not,adr, program, for, end_for, while, do, end_while, repeat, end_repeat, until, to, by, else, elsif
	},
	otherkeywords={
		:, :=, <>,;,\,.,\[,\],\^, TRUE, FALSE, \{attribute,  \'hide\'\}
	},
	keywords=[1]{
		case,of,if,then,end_if,end_case,super,function_block,extends,var,
		constant, byte,,end_var,var_input, real,bool,var_output,
		dint,udint,word,dword,array, of,uint,not,adr, :, :=, <>,;,\,.,\[,\],\^,program, function, for, end_for, while, do, end_while, repeat, end_repeat, until, to, by, else, elsif, sizeof
	},
	keywordstyle=[1]\color{blue},
	keywords=[2]{
		TRUE, FALSE
	},
	keywordstyle=[2]\color{codepurple},
	keywords=[3]{
		\{attribute,  \'hide\'\}
	},
	keywordstyle=[3]\color{codegray},
	sensitive=false,
	morecomment=[l]{//}, 
	morecomment=[s]{(*}{*)},
	morestring=[b]"
	morestring=[b]'
}


\title[ICSML: Industrial Control Systems Machine Learning Inference Framework]{ICSML: Industrial Control Systems ML Framework for native inference using IEC 61131-3 code}


\author{Constantine Doumanidis}
\affiliation{%
  \institution{New York University Abu Dhabi}
  \city{Abu Dhabi}
  \country{UAE}
}
\email{constantine.doumanidis@nyu.edu}

\author{Prashant Hari Narayan Rajput}
\affiliation{%
  \institution{NYU Tandon School of Engineering}
  \city{Brooklyn}
  \state{New York}
  \country{USA}
}
\email{prashanthrajput@nyu.edu}

\author{Michail Maniatakos}
\affiliation{%
  \institution{New York University Abu Dhabi}
  \city{Abu Dhabi}
  \country{UAE}
}
\email{michail.maniatakos@nyu.edu}


\begin{abstract}
  Industrial Control Systems (ICS) have played a catalytic role in enabling the 4th Industrial Revolution. ICS devices like Programmable Logic Controllers (PLCs), automate, monitor, and control critical processes in industrial, energy, and commercial environments. The convergence of traditional Operational Technology (OT) with Information Technology (IT) has opened a new and unique threat landscape. This has inspired defense research that focuses heavily on Machine Learning (ML) based anomaly detection methods that run on external IT hardware, which means an increase in costs and the further expansion of the threat landscape. To remove this requirement, we introduce the ICS machine learning inference framework (ICSML) which enables executing ML model inference natively on the PLC. ICSML is implemented in IEC 61131-3 code and provides several optimizations to bypass the limitations imposed by the domain-specific languages. Therefore, it works \emph{on every PLC without the need for vendor support}. ICSML provides a complete set of components for creating full ML models similarly to established ML frameworks.
  We run a series of benchmarks studying memory and performance, and compare our solution to the TFLite inference framework. At the same time, we develop domain-specific model optimizations to improve the efficiency of ICSML. To demonstrate the abilities of ICSML, we evaluate a case study of a real defense for process-aware attacks targeting a desalination plant.
\end{abstract}

\ccsdesc[500]{Computer systems organization~Embedded systems}
\ccsdesc[500]{Computing methodologies~Machine learning approaches}
\ccsdesc[300]{Security and privacy~Intrusion detection systems}

\keywords{industrial control systems, machine learning, framework, anomaly detection}

\maketitle

\section{Introduction}

As a term, Industrial Control Systems (ICS) encompasses all the devices that enable automation of an industrial process, including control systems and their instrumentation, network, and other systems. These ICS devices are frequently utilized in critical infrastructure to control critical physical processes. They ensure the smooth and reliable operation of industrial, energy, and commercial environments such as assembly lines, desalination plants, smart energy grids, chemical processing stations, and mining infrastructure. According to a 2021 report by Research and Markets \cite{research-markets-report}, the global market value for industrial controls is expected to grow to reach 49.6 billion US dollars by 2025.

With the advent of Industry 4.0, industries have started leveraging Information Technology (IT) devices in the Operational Technology (OT) sector, enabling real-time yield optimization, remote monitoring and control, predictive maintenance, increasing automation, and much more. The OT sector was once thought to be secure due to its air-gapped network. However, Industry 4.0 has blurred the boundaries between the IT and the OT sectors, resulting in the latter becoming susceptible to network-based remote attacks. For instance, the recent Log4j vulnerability (\texttt{CVE-2021-44228}), considered by many to be a vulnerability impacting the IT domain, has, in fact, even impacted ICS vendors. It has impacted more than 100 products from Siemens, 8 devices from Rockwell such as Plex IIoT and more, the Smart Script labeling software from WAGO, and many other manufacturers~\cite{misc:security-week}. As a result, adversaries can utilize vulnerabilities in the IT infrastructure to propagate onto the OT network and launch attacks with devastating impacts on critical infrastructure.

With the increase of internet-exposed OT devices and incentives such as rising payouts and a potential for severe damage due to abundant vulnerabilities, OT is becoming an increasingly attractive attack target. According to Packetlabs, globally, a total of 33.8\% of ICS computers were attacked in the first half of 2021. Moreover, the vulnerabilities for ICS grew by 41\%, compared to the last half of 2020, with 71\% of these vulnerabilities being remotely exploitable~\cite{misc:packetlabs}.

There have been many high-profile attacks in the ICS domain: Stuxnet infected over 200,000 computers and damaged a significant amount of centrifuges used for uranium enrichment~\cite{art:stuxnet}. The Ukraine power grid attack in 2015 began with stolen VPN credentials and saw adversaries opening up breakers, taking various substations offline, and overwriting legitimate firmware to disable remote commands. This led to a blackout affecting around 230,000 people~\cite{art:ukraine}. Other attacks have been thwarted; for instance, the attack on a water treatment plant in Oldsmar, Florida, where adversaries attempted to boost the level of sodium hydroxide in the water supply to 100x of its regular quantity~\cite{misc:florida}.

Attacks on ICS have been explored in literature. These include attacks on the Programmable Logic Controller (PLC) to change the control logic or its process parameters~\cite{art:icsref, proc:sabot, proc:clik}, modifying the firmware~\cite{proc:malware_physics}, or spoofing sensor readings as part of false data injection (FDI) attacks that gather and relay data to the PLC~\cite{art:fdi_grids, proc:fdi_control, art:fdi_power, proc:msf_desalination} for attacking the controlled physical process. Proactively uncovering vulnerabilities in PLC programming~\cite{icsfuzz}, patching~\cite{rajput2022icspatch}, and detecting the presence of malware~\cite{proc:RajputJtagMalware, rajput2021remote} are also topics of interest to the ICS community. Defenses for attacks on ICS hardware have also been studied extensively. Among others, these include protecting the PLC using control invariants (correlation between sensor readings and PLC commands)~\cite{proc:slueth}, extracting control logic rules~\cite{art:shadowplc} and detecting safety violations~\cite{proc:process_code_verifier}. Similarly, Machine Learning-based (ML) solutions~\cite{proc:ml_sol1, art:ml_sol2, proc:ml_sol3, art:ml_sol4, proc:ml_sol5} have been employed for detecting FDI attacks.

Moreover, the use cases for ML in the ICS domain are not limited to just attack detection. Its integration can immensely benefit machine vision for sorting, logistics, packing, predictive maintenance by monitoring machine status to predict hardware faults and schedule repairs, model optimization for auto-tuning control strategies by manipulating controller parameters, fault tolerance, and error detection~\cite{misc:ml-ics}.

\begin{figure}[t]
     \centering
     \subfloat[][Traditional ML-based defense for ICS, where sensor inputs have to be delivered to a regular IT device connected to the OT network.]{\includegraphics[trim={1.5cm 12cm 23cm 3cm}, clip, width=0.95\columnwidth]{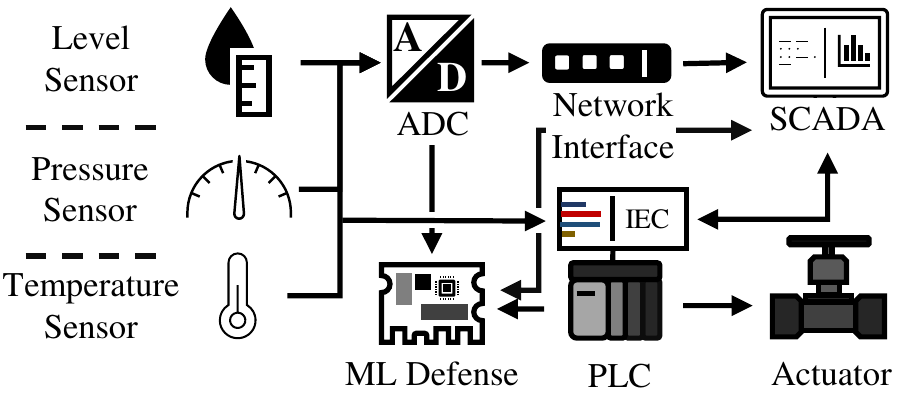}\label{fig:traditional_setup}}\\
     \subfloat[][ICSML setup, where the ML-based defense is implemented natively using standard PLC programming languages, removing the need for an external device and extra communication channels.]{\includegraphics[trim={1.5cm 7cm 23cm 7.5cm}, clip, width=0.95\columnwidth]{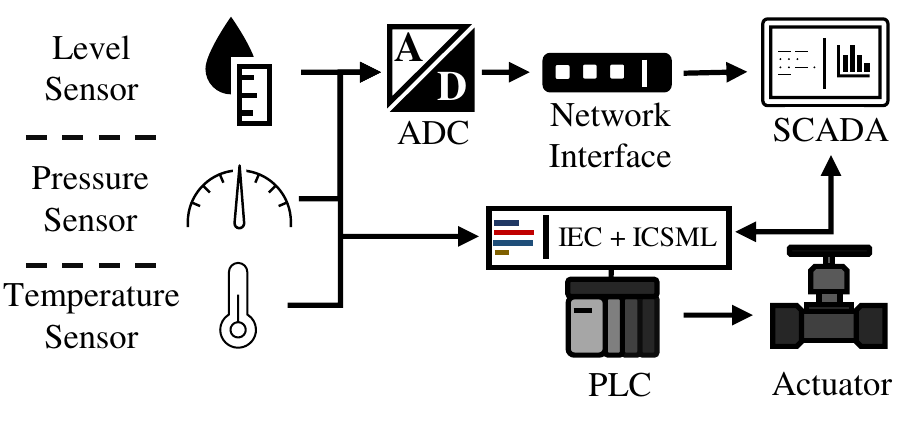}\label{fig:icsml_setup}}
     \caption{Traditional setup vs ICSML setup.}
     \label{fig:comparison}
\end{figure}

Such ML-based solutions are often built with high-level languages \cite{xu2021detecting, kim2019anomaly, zhang-fdi} and can only be executed on PLCs that employ an operating system (OS) that supports the execution of general-purpose binaries. Unfortunately, many of these PLCs are bare-metal or utilize proprietary OSes, forbidding such operations. Furthermore, these solutions often operate out-of-the-device either in online or offline data collection mode. Therefore, they cannot be realistically deployed due to the requirement of an additional device for each PLC with access to the sensor inputs used by the ML-based solution. Fig. \ref{fig:traditional_setup} shows the current state-of-the-art: Any ML-based solution has to receive the process inputs (typically analog) separately from the PLC, perform inference and potentially detect an anomaly. Also, this external device (typically a regular computer) must connect to the OT network to transmit its inference outcome. Apparently, a traditional IT device in the OT network can expand the attack surface and damage capabilities.

In order to remove the requirement of including and integrating a regular computer for ML-based security defenses, in this work, we present ICSML; an ML inference framework implemented \emph{natively} in Structured Text (ST), thus compatible with all IEC 61131-3 compliant devices. Since the framework is built in ST, it integrates with any IEC project and enables on-the-device implementation of ML-based defenses. Such a solution can work on the PLC without requiring additional external devices and remains tightly coupled with the sensor inputs relayed to the PLC. ICSML, displayed in Fig. \ref{fig:icsml_setup}, allows inference \emph{within} the PLC without 1) sacrificing inference accuracy, 2) affecting the PLC regular operation, and 3) requiring vendor support. Our contributions can be summarized as follows:

\begin{itemize}
    \item We formulate the problem of efficiently performing inference on PLCs using IEC 61131-3 languages by studying the expressivity limitations of the available languages, considering PLC hardware limitations, and factoring in the real-time constraints imposed by ICS operation environments.
    
    \item We develop a machine learning inference framework that works around IEC 61131-3 domain-specific language limitations and offers a complete set of ML components.
    
    \item We present an end-to-end methodology for performing ML inference on PLCs, including dataset formation, model building and training on established machine learning frameworks, model porting to ICSML, and execution on the PLC. 

    \item We evaluate the performance of the ICSML framework throu\-gh a series of benchmarks that highlight memory requirements and performance scaling with respect to model size, as well as a series of PLC-specific optimizations.
    
    \item We demonstrate the efficiency and non-intrusive nature of ICSML via a case study that implements a real on-PLC defense for a Multi-Stage Flash (MSF) desalination plant. 
    
\end{itemize}

\section{Preliminaries}

\subsection{Industrial Control Systems}
ICS encompass control systems and associated instrumentation that monitor, manage, control, and automate physical industrial processes, often employed in critical infrastructure. A fundamental building block of ICS are PLCs, which are integrated devices with ruggedized packaging designed for reliable operation in harsh industrial environments. PLCs work under a cyclical model of operation, known as the scan cycle, in which they read sensor inputs, perform computations based on the control logic and regulate the actuators. Programming the control logic of PLC devices is done by process engineers using a family of domain-specific programming languages defined by the IEC 61131 standard \cite{iec-tiegelkamp}.

\subsection{IEC 61131-3 Languages}\label{sec:languages}
IEC 61131 is an IEC standard in for programmable controllers. The third part of the standard (IEC 61131-3) defines several programming languages that include~\cite{proc:iec_languages}:

\noindent \textbf{Ladder Diagram (LD).} This is a low-level graphical language representing a ladder-like structure, consisting of vertical lines, the rails, connected by horizontal circuits. Here, input and outputs are represented by contact and coil symbols, respectively. LD is more suitable for maintenance technicians lacking familiarity with programming languages.

\noindent \textbf{Function Block Diagram (FBD).} A graphical language consisting of Function Blocks (FB), written in any IEC 61131-3 language, connected by horizontal lines known as the signal flow lines.

\noindent \textbf{Sequential Function Chart (SFC).} A graphical language specifying sequential behavior capable of integrating with other IEC languages. Rectangular blocks specify the phases of the control process and connect through flow lines.

\noindent \textbf{Structured Text (ST).} A high-level textual language that looks similar to the Pascal programming language. It allows programming complex functionalities to meet the growing sophistication of PLCs.

IEC projects consist of software blocks called Program Organization Units (POUs). These can be programmed using one IEC language and then utilized by other IEC languages, which makes IEC languages interoperable. POUs include Functions, which are defined similarly to functions found in general-purpose programming languages. Function Blocks are another fundamental POU and act similar to objects found in object oriented programming since they contain variables and methods that encapsulate state and functionality. IEC 61131-3 Interfaces act similarly to classes for Function Blocks that implement them. Finally, the standard also supports Structs that are defined alike C structures.

Functions and FB methods feature a special static part for the declaration and initialization of variables. Variables defined in the \texttt{VAR\_INPUT} and \texttt{VAR\_OUT} sections designate input arguments and return values respectively, while variables in the \texttt{VAR} section are considered local in scope. Finally, \texttt{VAR\_IN\_OUT} variables are used for both input and output.

\section{Problem Formulation}
In this work, we explore the ability of PLCs to \emph{efficiently} perform inference using IEC 61131-3 languages in order to reap the benefits of ML inference natively on the PLC. ML inference natively on the PLC is challenging since:

\begin{enumerate}\itemsep0em 
    \item PLC applications are written in limited expressivity domain-specific programming languages and compiled to non standard binaries. These languages have limitations such as no dynamic memory management, a lack of recursion support, and more that require novel approaches to support ML inference properly. This is elaborated further in Section \ref{sec:limitations}.
    \item PLCs come at a variety of computational capability configurations, with the majority incorporating a limited amount of memory and embedded processors, as further elaborated in Section \ref{ss:plchwlimitations}. Therefore, developed methods have to be both lightweight and not interfere with the primary computational objective of the PLC: control and monitoring of the physical process.
    \item Industrial processes typically have real-time requirements, with the scan cycle (sense-compute-actuate loop) potentially having periods in the order of milliseconds. Therefore, inference has to be carefully developed concerning the scan cycle, as further discussed in Section \ref{ss:realtimeconstr}.
\end{enumerate}

\subsection{IEC 61131-3 Languages Limitations}
\label{sec:limitations}
Implementing ICS machine learning solutions and all related functionality in the PLC vendor stack using IEC 61131-3 languages brings challenges that stem from their limitations due to their domain-specific nature. We consider the advantages and disadvantages of all available IEC languages and deem Structured Text to be the most well-suited language for developing ML applications as it resembles traditional programming languages and is the most versatile language defined in the IEC 61131-3 standard. Some significant limitations of the ST programming language are the following:

\noindent \textbf{Lack of Dynamic Memory Management:} ST, like the rest of IEC 61131-3 languages, lacks the ability to manage memory during run time dynamically. Among other things, this prevents the allocation of arrays with variable lengths.

\noindent \textbf{Inputs are ``Call-by-Value'':} Inputs to POUs declared under \linebreak\texttt{VAR\_INPUT} are passed by value to the called function, meaning copies of the input data are automatically created specifically for every POU call.

\noindent \textbf{Functions cannot call Function Blocks:} ST allows passing Function Blocks as arguments to Functions, but this is only to access them as data structures and not calling their methods.

\noindent \textbf{No Recursion:} Recursive calls of POUs, direct or indirect, in IEC 61131-3 languages are strictly forbidden because they do not allow calculating the maximum required program memory \cite{iec-tiegelkamp}. This limitation expands on instances of Function Blocks of the same type calling each other. Static compiler checks can be worked around by using methods like indirect recursion in which a POU of one type calls an intermediate POU, which then, in turn, calls another POU of the first type. However, these methods result in crashes during execution.

\noindent \textbf{No First-Class Functions:} Functions in ST are not considered first-class types and, as such, cannot be passed as arguments to POUs. This limitation prevents the employment of Functional Programming (FP) paradigms when programming in ST. However, these paradigms are often used in established ML frameworks to form data preprocessing pipelines and to configure components like the Keras lambda layer, which allows wrapping arbitrary expressions that map inputs to outputs as layers.

\begin{table*}[t]
\caption{PLC hardware specifications grouped by manufacturer. Time per instruction is measured by manufacturers using Load (LD), Floating Point (FP), Boolean (Bool) or a mix of instruction types. 
}
\footnotesize
\centering
\begin{adjustbox}{width=1\textwidth}
\begin{tabular}{llll} 
\toprule
\textbf{Manufacturer} & \textbf{Models}                            & \textbf{Average Time/Instruction ($\mu s$)} & \textbf{Memory / RAM}                               \\ 
\midrule
ABB                   & AC500 PM57x/58x/59x/595/50xx/55x/56x       & FP:0.7/0.5/0.004/0.001/0.6/1200             & 128-512KB/512KB-1MB/2-4MB/16MB/256KB-1MB/128-512KB  \\
Allen Bradley         & Micro 810/20/30/50/70,~CL 5380, 5560/70/80 & 2.5/0.3/0.3/0.3/0.3, N/A, N/A               & 2/20/8-20/20/40KB, 600KB-10MB, 3-40/2-32/2-32MB     \\
Delta Electronics     & AS300, AH500                               & 1.6, 0.02 LD                                & N/A, 128KB-4MB                                      \\
Eaton                 & XC152, XC300                               & N/A, N/A                                    & 64MB, 512MB                                         \\
Emerson               & Micro CPUE05/001, RX3i CPE400/CPL410       & 0.8 Bool/1.8, N/A                           & 64/34KB, 64MB/2GB                                   \\
Fatek                 & B1, B1z                                    & 0.33, 0.33                                  & 31KB, 15KB                                          \\
Festo                 & CECC-D/LK/S                                & N/A                                         & 16/16/44MB                                          \\
Fuji Electric         & SPH5000M/H/D/3000D/300/2000/200            & FP:0.0253/0.066/0.088/0.08/0.27/5600        & 4/4/2/2/2MB/128KB                                   \\
Hitachi               & Micro EHV+, HX, EHV+                       & N/A, 0.006 FP, 0.08                         & 1MB, 16MB, 2MB                                      \\
Honeywell             & ControlEdge R170 PLC~ ~ ~ ~                & N/A                                         & 256MB ECC                                           \\
Mitsubishi Electric   & MELSEC iQ-R/Q/L                            & 0.0098 FP/0.0016 LD/0.065 LD                & 4MB/64-896KB/64K Steps                              \\
Panasonic             & FP 7/2SH/0R/X0/0H                          & 0.011/0.03/0.08-0.58/0.08-0.58/0.01         & 1MB/20KB/64KB/16KB/64K Steps                        \\
Rexroth (Bosch)       & XM21/22/42, VPB                            & FP:0.026/0.013/0.02/0.02                    & 0.5/0.5/2/16GB                                      \\
Schneider Electric    & Modicon M221/241/251/262                   & 0.3/0.3/0.022/0.005                         & 256KB/64MB/64MB/32MB                                \\
SIEMENS               & SIMATIC S7-1200/1500                       & 2.3/0.006-0.384                             & 150KB/150KB-4MB                                     \\
WAGO                  & PFC100/200                                 & N/A, N/A                                    & 256/512MB                                           \\
\bottomrule
\end{tabular}
\end{adjustbox}
\normalsize
\label{tbl:plc_hardware}
\end{table*}

\subsection{PLC Hardware Limitations}\label{ss:plchwlimitations}
Given that PLCs are often deployed in interconnected harsh industrial environments, their design usually prioritizes robustness, efficiency, connectivity, and ruggedness over the inclusion of high-performance computational hardware. 

As can be seen in Table \ref{tbl:plc_hardware}, computationally entry-level PLCs like Allen Bradley Micro 810 utilize low power CPUs and feature a limited amount of memory (just 2 KB). Mid-tier performance-wise PLCs, like the Schneider Electric Modicon M241, have faster multi-core processors and upgraded RAM (around 64 MB). Finally, higher-end PLCs, like the WAGO PFC 200, typically house faster ARM-based CPUs and are paired with 512 MB of RAM.

Considering the limited resources available on PLCs, ICS machine learning applications must be written to make efficient use of I/O, CPU time, and memory to seamlessly operate and achieve the desired performance metrics without resource starving other ICS tasks running on the PLC.

\subsection{Real Time Constraints}\label{ss:realtimeconstr}

ICS processes typically involve tasks that run cyclically, such as PID feedback loops. As such, PLCs operate by following a periodic model based on the scan-cycle sequence. At the beginning of the scan cycle, the PLC captures the values from its inputs, which are usually sensor readings, and loads them into memory for computations. After, the PLC executes the instruction sequence that it was programmed to do using IEC 61131-3 languages. Finally, based on the results of the previous step, the PLC outputs are updated. These outputs can be utilized for various purposes such as actuator activation and control, data collection, etc.

The desired length of the scan cycle can vary depending on the ICS environment. For example, PLCs used to control robots in industrial manufacturing might be required to respond within milliseconds, especially when human safety is involved. On the other hand, slower petrochemical processes can even be controlled using second-long PLC scan cycles.

The number of calculations performed in a single PLC scan cycle is limited by the length of the scan cycle and the computational power of the deployed PLC. Violation of ICS process real-time constraints set by the scan cycle, can have disastrous ramifications.

\section{ICSML Framework}
ICSML is a Machine Learning Inference Framework for ICS environments built using Structured Text. Its goal is to provide the ML application engineer with a code base and structured approach for effortlessly building and deploying device-agnostic, real-time, and efficient ML solutions on PLC hardware. As discussed in Section \ref{sec:languages}, thanks to IEC 61131-3 language interoperability, process engineers can use all IEC languages to take advantage of ICSML functionality.

\subsection{Framework Architecture}
ICSML is comprised of Activation Functions, Math and Utility Functions, Data Structures, Layers, and Models. These components have been modeled and developed similarly to popular machine learning frameworks in order to ensure compatibility and enable users to port their models to ICSML easily.

\noindent \textbf{Activation Functions:} One of the fundamental components of an Artificial Neural Network (ANN) neuron is its activation function which is applied to the weighted sum of its inputs to calculate the output. Activation functions are used to introduce non-linearity, allowing ANNs to achieve more complicated goals using fewer neurons. ICSML provides parameterizable implementations for the Binary Step, Exponential Linear Unit, Rectified Linear Unit (ReLU), Leaky ReLU, Sigmoid, Softmax, Swish, and Hyperbolic Tangent (Tanh) activation functions.

\noindent \textbf{Math \& Utility Functions:} Machine learning inference involves a series of matrix and vector multiplications that rely on the dot product operation, which is implemented as a function in ICSML. Beyond this, ICSML provides utility functions \texttt{BINARR} and \texttt{ARRBIN} that provide abstractions to load and save array data from and to binary files. Among other things, these can be used to form datasets, load model weights and biases, and log inference results.

\noindent \textbf{Layers \& Data Structures:} Like in other ML frameworks, layers are the fundamental building blocks of ML models in ICSML. Dense layers feature a number of neurons with corresponding weights and biases used to calculate the layer outputs. Concatenation layers combine their inputs and can be used to build ML models with parallel sections that branch out and merge. Activation layers apply an activation function to their inputs. Finally, the ICSML framework offers a convenient way to manage layer memory by providing the user with the \texttt{dataMem} structure which associates memory areas with their metadata.

\noindent \textbf{Models:} Following other frameworks, ICSML Models consist of an array of layers wired together, and an inference method. Among others, the framework enables building traditional densely connected feedforward ANNs, Convolutional Neural Networks (CNNs), branching Residual Neural Networks (ResNets), and even more sophisticated architectures such as Recurrent Neural Networks (RNNs).

\subsection{ICSML domain-specific optimizations}
Enabling efficient ML inference in ICSML requires making domain-specific optimizations and architectural design decisions that consider factors unique to the PLC ecosystem. 

\subsubsection{Memory Management Abstractions}
Layered ML models require allocating memory buffers for weights matrices, biases vectors, and activations for each model layer. As discussed in Section \ref{sec:limitations}, ST has several nuances pertaining to memory management. ICSML disburdens the programmer from manually managing memory by offering certain abstractions, involving the structured declaration of layer sizes via constants and then using them for the static allocation of memory areas. Pointers to these memory areas are then associated with data dimensionality information and metadata via the \texttt{dataMem} structure. This abstraction allows the programmer to easily manage memory from a higher level using a single object and removes the need for manually handling layer dependencies brought on by the lack of dynamic memory management. The ICSML memory management structure also hides the data required by framework internals to function and works around the data duplication issue that occurs when passing arrays by value to function calls by declaring them under \texttt{VAR\_INPUT}. The latter is especially important for memory limited PLCs like the Mitsubishi MELSEC iQ-R, where evaluating a 512 neuron dense layer during inference by passing the weights and biases arguments ($\approx$ 2MB) as \texttt{VAR\_INPUT} might overflow its 4 MB memory. While these arguments can be passed by reference using \texttt{VAR\_IN\_OUT} declarations, this data would no longer be accessible externally in a structured way by the programmer when using the dot accessor on the layer. In this sense, the ICSML memory management mechanism combines the benefits of both \texttt{VAR\_INPUT} and \texttt{VAR\_IN\_OUT} declarations by minimizing memory duplication while permitting data accessibility. The ICSML memory management system can also work around ST limitations when implementing other algorithms like convolution, BLAS, divide and conquer style, and others.

\subsubsection{Code Templates}

ML engineers often build models that include components and constructs which implement custom functionality beyond the one found in standard layers. Such components include lambda and masking layers, and data preprocessing pipelines. Traditional ML frameworks often support this through language functionality, like functional programming paradigms, that is not present in ST. For example, something similar to the Keras lambda layer, which enables converting arbitrary expressions into ML layers, can be built in IEC 61131-3 languages by emulating FP style programming using function blocks as closures. However, the amount of redundant code and programming effort required to implement this renders this method inconvenient. ICSML avoids employing such emulation tactics by instead including ST interface templates for all core components. These can be implemented by ML engineers to effortlessly and optimally introduce custom functionality in their models while minimizing boilerplate code.

\subsubsection{Non-Chained Function Calling}

ML models built with ICSML are represented in ST using objects that encapsulate the model architecture, metadata, and inference logic.
Layers inside model FBs are interconnected by sharing \texttt{dataMems}, and are evaluated by the model evaluation function, which linearly calls their respective evaluation functions. This linear method of performing inference is more efficient than implementing ML inference using layer objects linked via references and chained function calls. While the latter would have been more straightforward and offered some additional convenience to the programmer, implementing the linear inference method is necessary to work around the recursion restrictions of the language.

\subsubsection{Self-Contained System Architecture}
An important decision that had to be made when designing the ICSML framework was the reliance on external code versus implementing all necessary functionality as part of the framework. ML frameworks rely extensively on computationally intense mathematical calculations, like matrix multiplication. Many PLC manufacturers provide optimized libraries for their ecosystems or allow loading external precompiled libraries that will support performing advanced mathematical operations optimally. However, utilizing readily made libraries by manufacturers limits cross-compatibility and often requires separate licensing. Furthermore, unlike in other programming environments, open source ICS libraries, like OSCAT \cite{OSCAT} are rare and do not cover the needs of an ML framework. Therefore, for ICSML, we chose to implement all the necessary functionality as part of the framework using ST in order to offer cross-platform compatibility with the whole IEC 61131-3 compliant ecosystem.

\subsection{Methodology for Porting an ML Model to ICSML}
\label{sec:model_porting}

\begin{figure}
     \centering
     \includegraphics[trim={0.75cm 7cm 17cm 4.5cm}, clip, width=0.98\columnwidth]{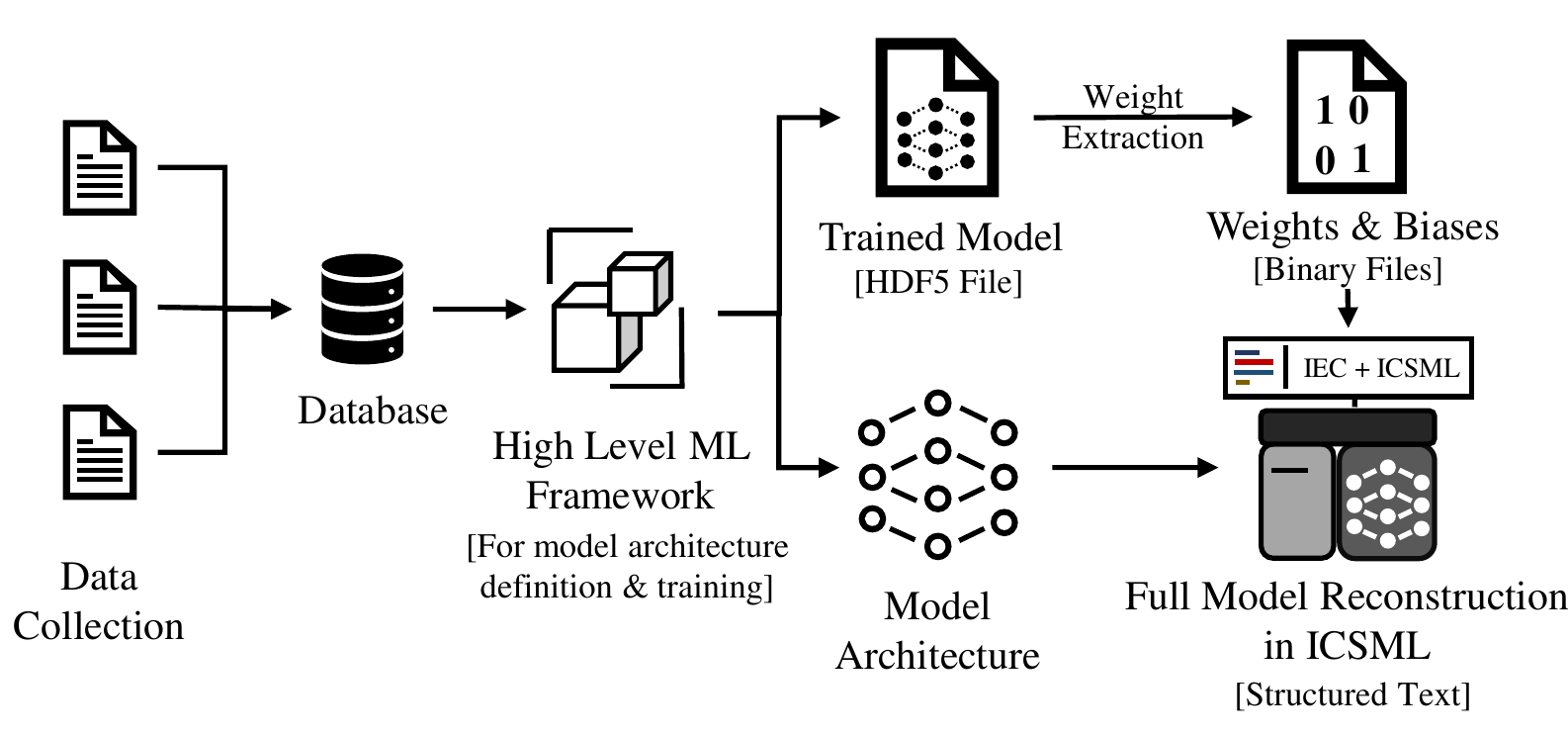}
     \vspace{-0.15in}
     \caption{ICSML model porting methodology overview.}
     \vspace{-0.1in}
     \label{fig:icsml_port_methodlogy}
\end{figure}

The ICSML framework was designed to enable ML application engineers to effortlessly design and train their models on popular ML frameworks, and run them on ICS platforms. This section discusses the steps involved in building and porting a machine learning model to ICSML. An overview of the process is depicted in Figure \ref{fig:icsml_port_methodlogy}.

Data collection is an important step in creating ICS ML models, as the quality of the formed dataset directly impacts the quality of the end model. Data collection can be done by recording PLC inputs using an external setup with an Analog to Digital Converter (ADC). However, it is recommended that this process take place directly on the PLC to account for potential PLC ADC effects like quantization noise and errors, thus ensuring that the collected samples match what is read by the PLC. Using the ICSML \texttt{ARRBIN} function, PLC inputs can be recorded to binary files.

\begin{lstlisting}[language=ST, frame=single]
ICSML.ARRBIN('PLC_Inputs.bin', Inputs_size * SIZEOF(REAL), ADR(Inputs));
\end{lstlisting}

Model architecture design and training can be done using established high-level ML frameworks like TensorFlow \cite{tensorflow2015-whitepaper}, Pytorch \cite{PyTorch}, and Caffe \cite{caffe}. Designing, training, and exporting the ML model with the purpose of then porting it to ICSML do not differ from standard procedure.

Weights and biases extraction from the exported ML model is necessary to reconstruct the trained model in ICSML. During this step, the exported model file is read, and its weights and biases are saved into binary files, which are then loaded onto the PLC.  \label{sec:extract_weights}
Given the knowledge of the trained model architecture and its extracted weights and biases, the model can be reconstructed in ICSML using a structured approach that first requires defining layer sizes as constant variables. 
Then, array variables for weights, biases, output memory buffers and dimensions of each layer are declared.

\begin{lstlisting}[language=ST, frame=single]
L1_weights: ARRAY[0.. L1_size * input_size - 1] OF REAL;
L1_biases: ARRAY[0.. L1_size - 1] OF REAL;
L1_buff: ARRAY[0.. L1_size - 1] OF REAL;
L1_dimensions: ARRAY[0..0] OF UINT := [L1_size];
\end{lstlisting}

After, memory buffers and size metadata are used to construct ICSML \texttt{dataMem} structures for each layer:

\begin{lstlisting}[language=ST, frame=single]
L1_dataMem: ICSML.dataMem := (address:=ADR(L1_buff), length:=L1_size , dimensions:=ADR(L1_dimensions), dimensions_num:=1);
\end{lstlisting}

Subsequently, layers are instantiated and placed into an array which is used to construct the model.

\begin{lstlisting}[language=ST, frame=single]
L1_layer: ICSML.Dense := (input:=input_layer.output , output:=L1_dataMem , weights:=ADR(L1_weights), biases:=ADR(L1_biases), activation:=ICSML.activationType.ReLU);
Model: ICSML.Sequential := (layers:=ADR(layers_array), layers_num:=UPPER_BOUND(layers_array, 1)+1);
\end{lstlisting}

Finally, weights and biases are loaded using \texttt{BINARR}, and the model is ready to be used for inference.

\begin{lstlisting}[language=ST, frame=single]
ICSML.BINARR('L1_biases.bin', L1_size * SIZEOF(REAL), ADR(L1_biases));
Model.evaluate();
\end{lstlisting}

\section{ICSML Benchmarking}
\label{sec:benchmarking}

Performance and efficiency are vital for running ML applications on PLCs since these devices have limited memory and CPU resources bound by real-time constraints. In this section, we measure the performance of the proposed framework when deployed on PLC hardware such as WAGO PFC100 (Single-Core, 600MHz ARM Cortex-A8, 256MB RAM) and a BeagleBone Black (Single-Core, 1GHz ARM Cortex-A8, 512MB RAM). For investigating the scalability of the framework, we study various configurations of ICSML components, their memory usage, and CPU times. 

Comparing ICSML to other state-of-the-art inference frameworks is challenging, as PLCs do not generally support loading code outside the vendor stack. In our study we utilize the BeagleBone Black which, while not a traditional PLC, the Codesys Runtime officially supports for development and prototyping purposes, and can also be used as a ``soft PLC''. Using it, we are able to compare ICSML to the TensorFlow Lite framework. When considering factors that impact performance, the BeagleBone Black and WAGO differ mainly in processor clock speed and the amount of available RAM. Using both devices for our testing gives us better insight into how the framework performs with respect to these two parameters and what kind of models PLC hardware can run.


\subsection{Memory Limitations}
\begin{figure}[t]
    \centering
    \includegraphics[angle=0,width=1.0\columnwidth]{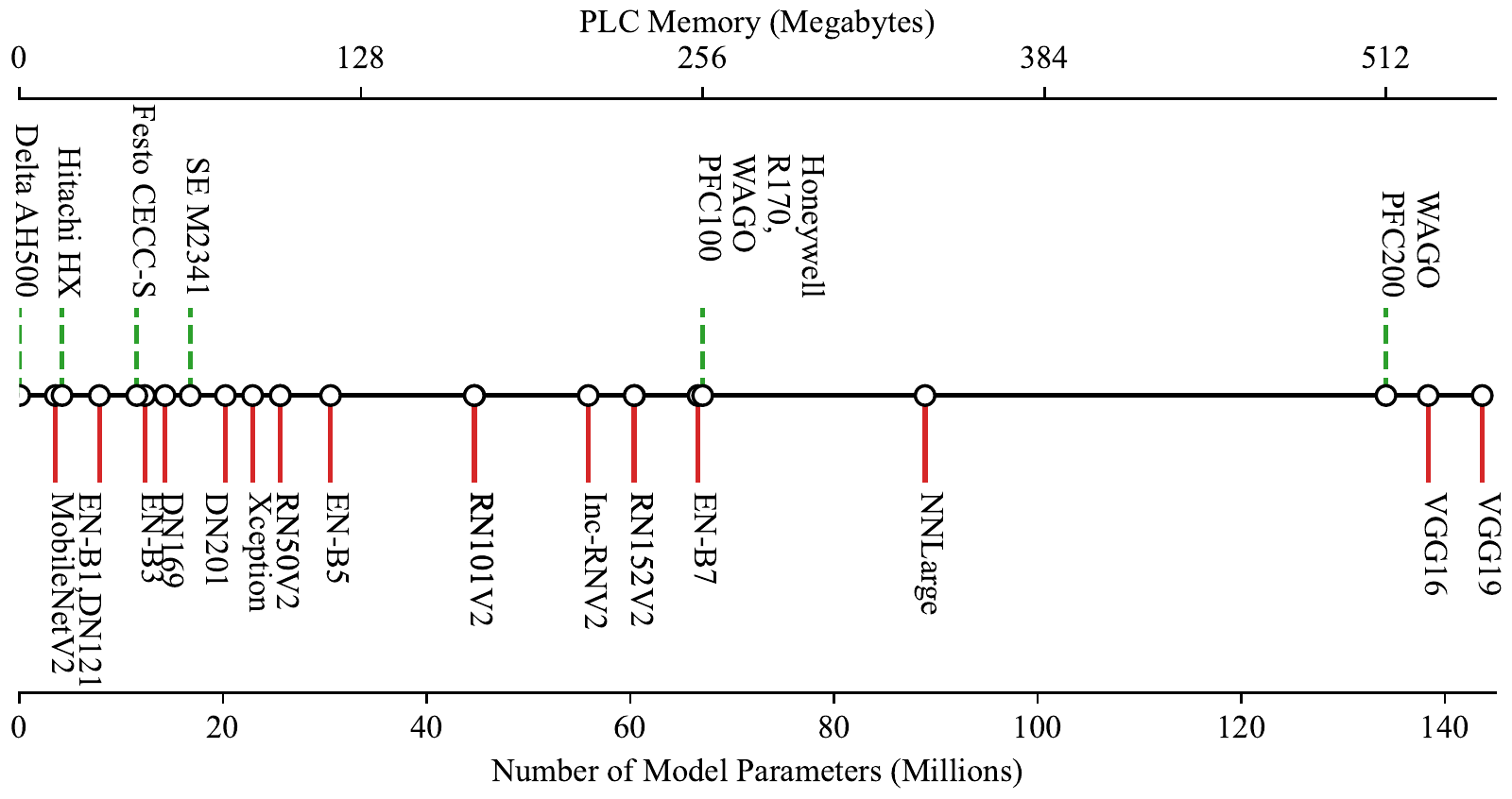}
    \captionof{figure}{Upper: PLCs and their memory in megabytes. Lower: DL models from the Keras library with their sizes in millions of parameters (32-bit). Abbreviations: DN - DenseNet, EN - EfficientNet, MN - MobileNet, NN - NASNet, RN - ResNet.}
    \label{fig:models_memory}
\end{figure}

We instantiate a test ICSML application with various configurations to study the memory limitations of PLC ML applications. Using the benchmarking tools, a linear relation is observed between layer sizes and memory usage since each layer on the PLC occupies memory equal to the sum of its weights matrix, biases vector, and the output memory buffer. To put the size of PLC memory into perspective with the size of ML models in the number of parameters and size occupied on disk, consider Figure \ref{fig:models_memory}. The figure contrasts popular Deep Learning models from the Keras Application library with various PLCs according to model size and memory availability. The figure shows that most presented PLCs can only run the smaller models. As previously discussed using Table \ref{tbl:plc_hardware}, PLCs typically come with limited size memory, making it imperative that ML models deployed on ICS hardware use memory efficiently.

\subsection{Layer Stacking Scaling}

To study the performance implications of adding additional dense layers to an ICSML model, we utilize a fully connected model with 64 32-bit input and output features. The input layer performs a simple copy operation, and its evaluation function takes 3 \textmu s to return. ICSML uses the \texttt{BINARR} and \texttt{ARRBIN} functions to load and store input and output vectors, which take approximately 396 \textmu s and 530 \textmu s of CPU time per call respectively on the BeagleBone Black, and 447 \textmu s and 535 \textmu s on the WAGO PFC100. Initially, we instantiate a single dense layer and add an additional 64 neuron layer with the ReLU activation in every iteration of the experiment.

\begin{figure}[t]
    \centering
    \includegraphics[width=1\columnwidth]{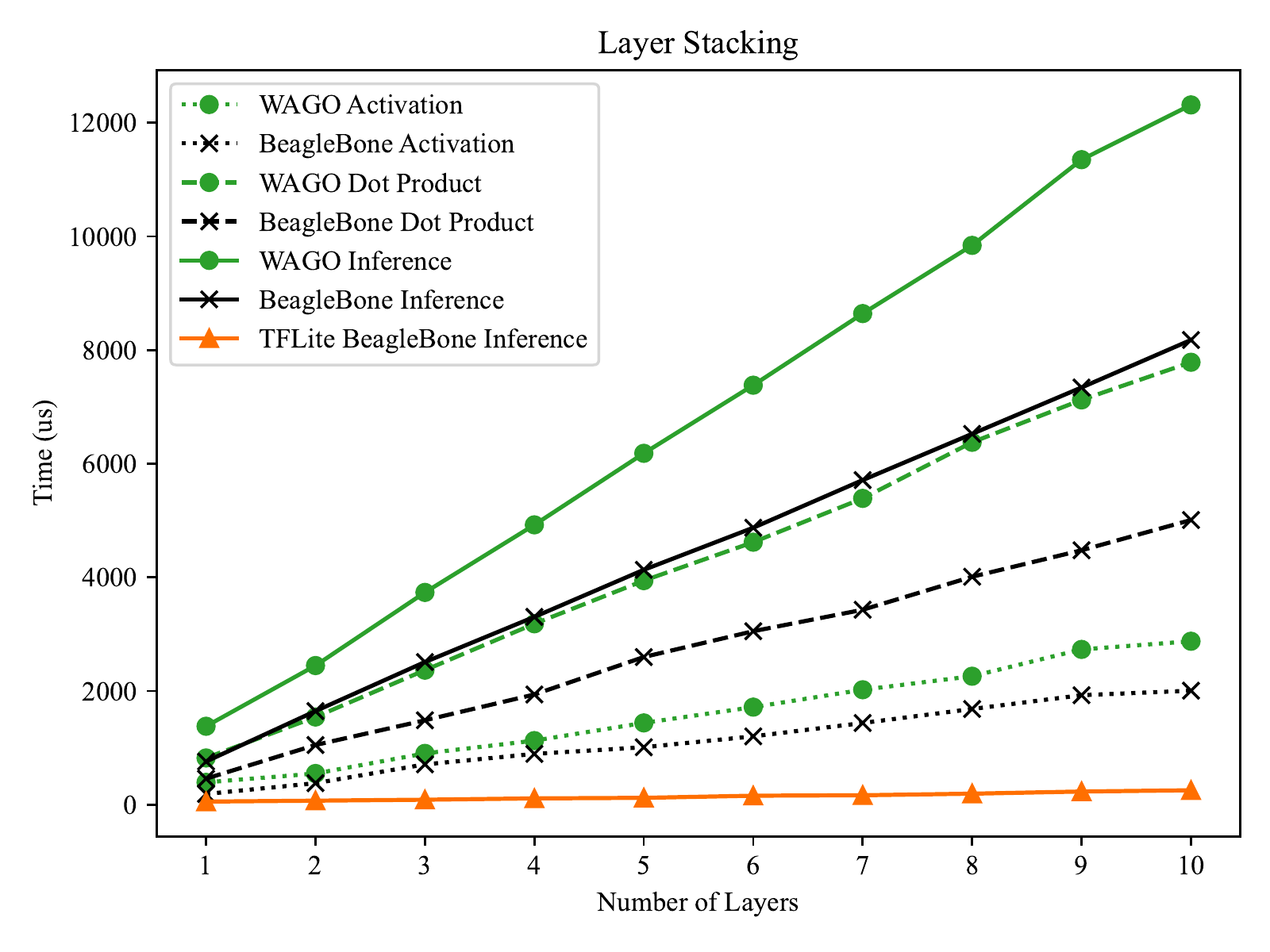}
    \vspace{-0.3in}
    \caption{Scaling of activation function, dot product and model inference CPU times for the number of model layers on a WAGO PFC100 and a BeagleBone Black running ICSML, and a BeagleBone Black running TF Lite.}
    \label{fig:layer_stacking}
    \vspace{-0.5cm}
\end{figure}

As can be observed in Figure \ref{fig:layer_stacking}, CPU time for the dot product operation, activation function application, and the model as a whole scale linearly. In the case of the BeagleBone Black, each additional layer in the test model adds approximately 455.186 \textmu s, 181.81 \textmu s, and 741.863 \textmu s to each of the aforementioned execution times respectively, while for the WAGO PFC100 these numbers increase to 696.435 \textmu s, 248.347 \textmu s and 1093.565 \textmu s. 

Comparing the total inference time of the ICSML benchmark model to that of an equivalent model running on the BeagleBone Black with the TensorFlow Lite inference framework shows that inference in TFLite is on average 29.38x and 44.69x faster than inference using ICSML running on the BeagleBone Black and the WAGO PFC100 respectively.

\subsection{Layer Size Scaling}
To understand the performance impact of layer widths in ICSML, we instantiate a simple model with a 32 features input layer and a single dense layer that also applies the ReLU activation function. On each iteration of the experiment, the number of neurons of the dense layer is doubled. Results exhibit a linear scaling similar to that in Figure \ref{fig:layer_stacking}. Execution time for the dot product operation, activation function application, and the model as a whole scales almost linearly with respect to the number of neurons. On average, a single neuron adds approximately 9.326 \textmu s to the total model inference time on the BeagleBone Black and 13.722 \textmu s on the WAGO PFC100. Comparison of execution time between an equivalent TFLite benchmarking model and the ICSML model running on the BeagleBone Black and the WAGO PFC100 shows that TFLite is faster by 20.78x and 30.68x, respectively.

\subsection{Understanding Performance}
To understand the performance of ICSML, and how it compares to Tensorflow Lite running on the BeagleBone Black ($\sim$20-30x faster), we run a series of experiments with our benchmarking tools, and ICSML code. First, ICSML performance is measured using the Codesys profiler, which inherently introduces instrumentation overhead. Our experiments show that ICSML code without profiler instrumentation ran approximately 2x faster. Furthermore, the ICS code compilation process prioritizes predictability over performance, consequently compiler optimizations are expected to be conservative, if any. We faithfully reimplemented ICSML in C++ and
found our code compiled with the \texttt{-O3} optimization flag ran $\sim$4x faster compared to the \texttt{-O0} version. Finally, the remaining $\sim$3x performance difference is due to the fact that, unlike TFLite, ICSML code has to work around domain specific limitations, and cannot leverage optimized open source libraries for operations such as matrix multiplication.

\section{PLC-specific Model Optimizations}

As previously discussed, PLC memory and processing capabilities are limited. These constraints combined with the predetermined length of the scan cycle used to govern the underlying ICS process delineate the time available for ML inference. In order to fit larger and more complex models on ICS hardware, certain model optimizations that allow for compression and inference latency reduction can be considered. These optimizations include weight quantization, pruning and clustering. The former two can offer both compression and latency reduction benefits, while clustering only offers compression benefits and its application to ICSML models does not differ from standard ML models.

\subsection{Quantization}

Integer Quantization \cite{dl-quant-eval} is a popular technique for ML model optimization that replaces floating point model parameters with integer representations. This allows for significant model compression, since parameters are represented with lower precision, and latency reduction during inference, due to integer arithmetic being typically faster than its floating point counterpart. The use of lower precision arithmetic in quantization can result in a loss in model accuracy, which, however, can be minimized to acceptable margins or even alleviated completely \cite{han2015learning, zhou2017incremental, zhang2018lq}.

For studying the effects of quantization on ML models running on PLCs, we port a trained 3-layer fully connected MNIST classification model to ICSML. We then isolate and quantize its second hidden layer, which accepts 512 inputs, and outputs 512 activations. We quantize the initial REAL (32-bit) floating point weights using the SINT (8-bit), INT (16-bit), and DINT (32-bit) IEC 61131-3 Integer types. As shown in Table \ref{tbl:quant_memory}, SINT and INT quantization decreases the memory requirements of the layer by 74.66\%, and 49.71\% respectively. DINT quantization does not allow for any model compression; however, it provides benefits in inference latency reduction. 

\begin{table}
\centering
\caption{Memory Requirements in bytes of a fully connected 512-neuron layer with 512 inputs for various quantization schemes. Scaling Factors and biases are REAL numbers. 
}
\begin{adjustbox}{width=1\columnwidth}
\begin{tabular}{lllll}
\toprule
Scheme                        & Weights   & Biases & Scaling Factors & Total      \\ 
\midrule
SINT (8-bit)                  & 262,144   & 2048   & 2052            & 266,244    \\
INT (16-bit)                  & 524,288   & 2048   & 2052            & 528,388    \\
DINT (32-bit)                 & 1,048,576 & 2048   & 2052            & 1,052,676  \\
REAL (32-bit) & 1,048,576 & 2048   & N/A             & 1,050,624  \\
\bottomrule
\end{tabular}
\end{adjustbox}
\label{tbl:quant_memory}
\end{table}

By analyzing the number and types of arithmetic operations performed during inference for the isolated layer, it can be shown that inference without quantization requires 262,144 floating point multiplications and 262,656 floating point additions. On the other hand, using integer quantization requires only 1024 floating point multiplications and 512 floating point additions but also requires 262,144 integer multiplications and 262,144 integer additions. 

\begin{figure}[t]
\centering
\includegraphics[trim=0.4cm 0.1cm 1.2cm 0.48cm, clip, width=1.0\columnwidth]{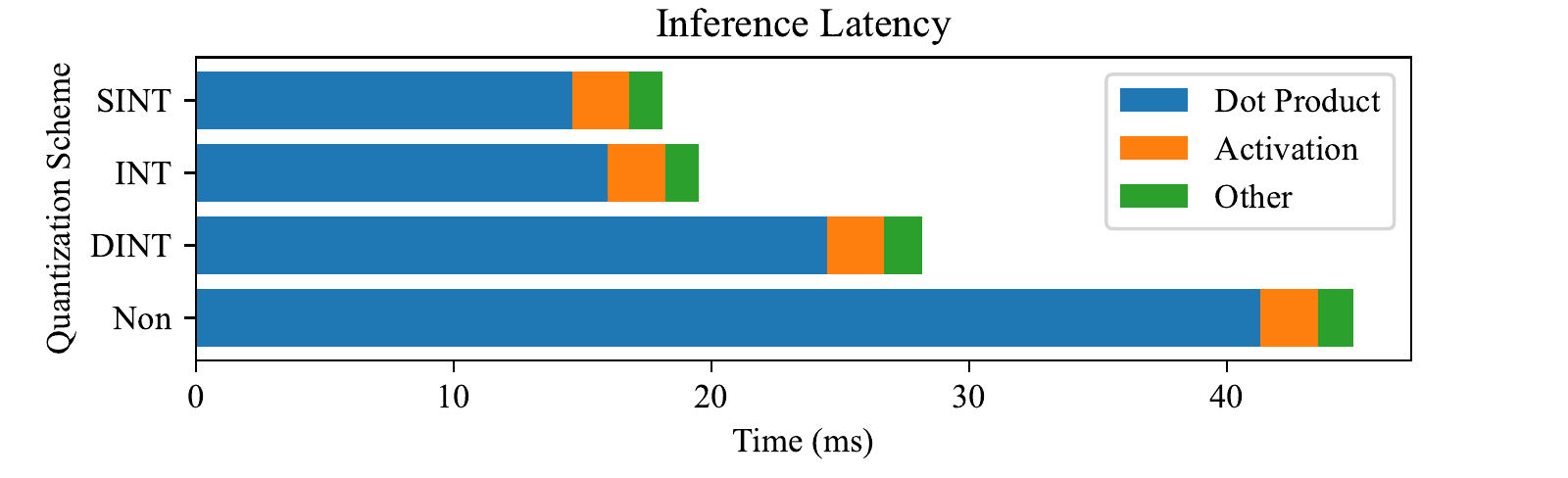}
\caption{Dense layer (512 inputs/outputs, ReLU activation) inference latency under various quantization schemes. The Other portion of inference latency includes dequantization and activation quantization time for the quantization setups.}
\label{fig:quant_performance}
\end{figure}

As shown in Figure \ref{fig:quant_performance}, SINT quantization results in 59.71\% decreased inference latency, while for INT and DINT quantization latency is decreased by 56.52\% and 37.23\% respectively. Quantization mainly influences the dot product portion of inference time, where most multiplication and addition operations occur. On the other hand, quantization does not influence activation time. Dequantization and activation dequantization time is negligible.

\subsection{Pruning}
Weight Pruning \cite{zhang2018systematic} is a widely used model optimization technique that promises significant model size reduction and latency time improvements by exploiting redundancy in model weights. Pruning forces weight sparsity during training by setting weights to zero to enable model compression and skipping arithmetic operations during inference.

Compression benefits of weight pruning differ on a case-by-case basis and are evident immediately after training by examining the model size on disk. This section does not focus on compression but investigates the potential of pruning for inference latency reduction.

As also shown in Figure \ref{fig:layer_stacking}, inference latency is linearly related to the number of calculations that are made when evaluating the model. Pruning reduces the number of calculations required since basic mathematical operations with a zero operand are redundant. To fully reap the inference benefits that pruning offers, standard fully connected layers can be reimplemented in ICSML on a case-by-case basis to skip redundant operations. 

However, even for models that do not use specially crafted layers, pruning can be potentially beneficial if there is underlying support from the runtime or hardware. To investigate this, we perform a series of experiments on the WAGO PFC100. The base of the experiment is a fully connected layer with 784 input features and 512 activations. The average dot product time for the layer using the original weights is 52.13 ms, and setting all weights to zero decreases this time to 47.62 ms. This minor reduction in latency does not indicate automatic operation skipping by the PLC runtime, so we test a manual implementation where an \texttt{IF} statement evaluates the weight being equal to zero to skip performing arithmetic operations. This results in an average latency of 50.84 ms. We repeat these experiments using SINT quantization, and the average latency for the three experiments are 36.39 ms, 35.69 ms, and 20.87 ms, respectively. Changing the operation-skipping statement to evaluate the equality of both inputs and weights to zero results in 34.19 ms average latency.

These results show that there is no underlying runtime or hardware supported speedup due to pruning since the dot product operation of a vector with a zero weight vector is not significantly faster. Manually skipping the operations by evaluating if the weight vector element is equal to zero seems to only be beneficial when combined with quantization since this check adds an overhead to calculation times when evaluating REAL floating point weights.

\subsection{Multipart Inference}
Short scan cycles, large models, and low computational power can make it impossible to fit ICS tasks and inference into a single scan cycle. To address this, ICSML enables performing multipart inference by splitting computations across multiple scan cycles. As an example, using multipart inference we were able to execute a model based on the MobileNet architecture ($a=0.25$, 4xConv2D, 7xBatchNorm+ReLU, 3xConvDW) on the BBB at a 90 ms scan cycle with an output latency of 1.17s.

\section{Case Study: ML-based anomaly detection  for a Desalination Plant}
This section presents a real on-PLC ML application built with ICSML. The case study implements a defense mechanism against process-aware attacks targeting a Multi-Stage Flash (MSF) desalination process.

\begin{figure}[t]
\centering
\includegraphics[clip, trim=6cm 1cm 3.8cm 2.5cm, width=0.9\columnwidth]{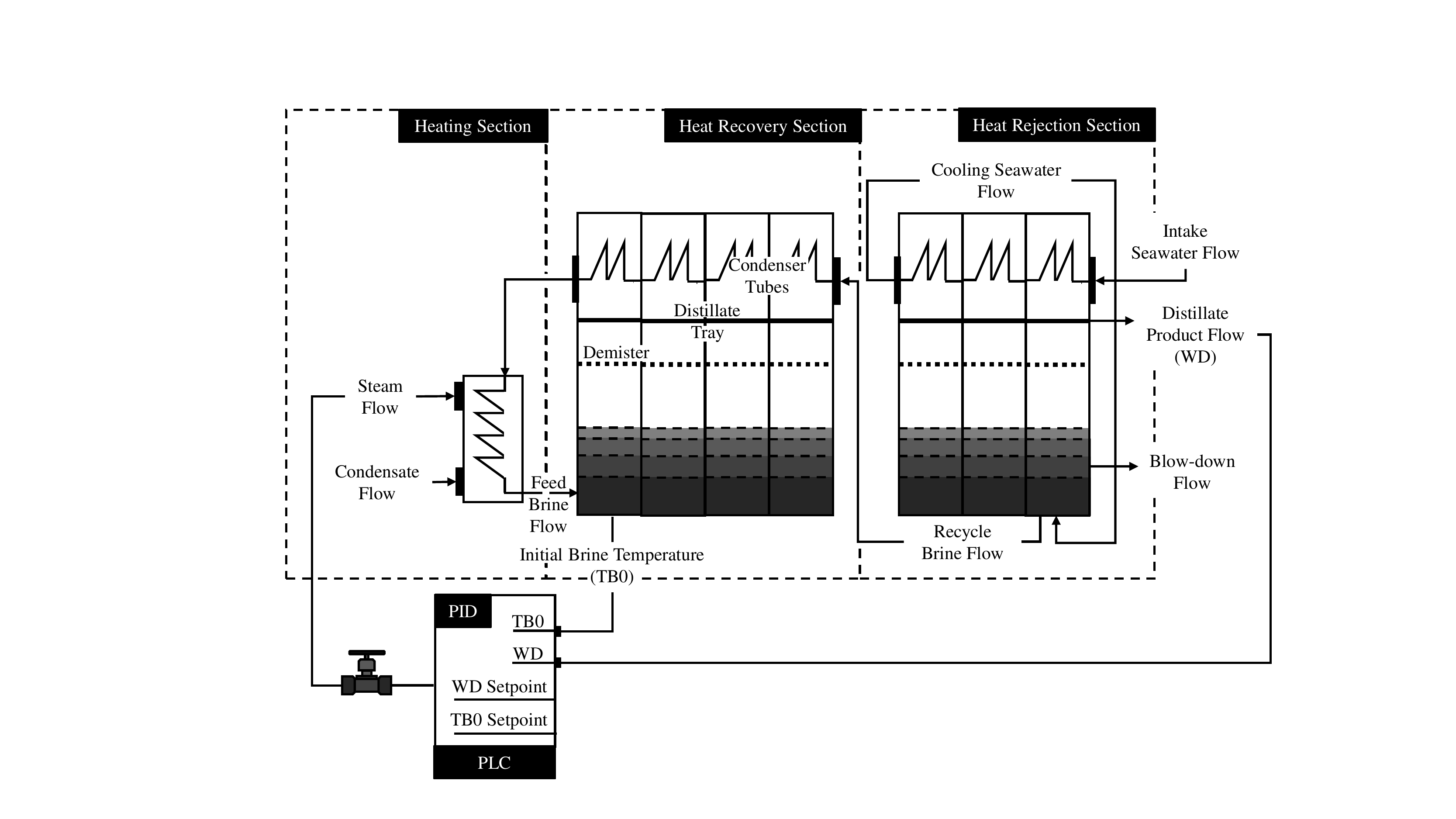}
\caption{Diagram of a Multi Stage Flash Desalination Plant.}
\label{fig:msf_plant}
\end{figure}

MSF desalination, shown in Figure \ref{fig:msf_plant}, is a widely used process that converts high salinity seawater into potable water. A process-aware attack on the plant assumes that the attacker has compromised the plant control system and has prior knowledge of the regulated physical process that can be exploited for inflicting damages. Such attacks directly impact the controlled process and can be detected by observing variables associated with the ICS system state. PLCs utilize these variables, that include sensor readings and actuator states, to control the dynamic process. Thus, it is possible to formulate process-aware attack detection as an anomaly detection problem that can be solved by employing machine learning methods running directly on PLC hardware.

For testing the applicability of ICSML in process-aware attack detection, we employ a simulated model~\cite{msf-model} of an MSF Desalination plant validated against actual data from Khubar II MSF plant in Saudi Arabia~\cite{proc:msf_desalination}. We connect the simulation in a Hardware-In-The-Loop (HITL) setup, where MATLAB Simulink simulates the core process, and a connected PLC controls part of the physical process by regulating the Steam Flow Rate (Ws).
In this setup, the PLC interfaces with the rest of the simulation by receiving Initial Brine Temperature (TB0) and Distillate Product Flow Rate (Wd) as inputs, and outputs the Ws control signal, which is calculated by a cascading PID controller setup.

Using the simulated setup we build a dataset which includes data collected under standard plant operating conditions and under a series of different simulated attacks that have been shown to inflict palpable damages \cite{proc:msf_desalination}. The dataset contains about 22 hours and 45 minutes of MSF plant operation data, out of which about 11 hours and 6 minutes are data collected under 7 different process-aware attacks that aim to influence different performance metrics. TB0 and Wd measurements observed by the PLC are collected and labeled accordingly at a 100ms interval to match the PLC scan cycle.

For detecting process-aware attacks, the PLC is considered to have access to ordered TB0 and Wd readings from the past 20 seconds. Given that the PLC scan cycle is 100ms, we design a densely connected classifier with 400 inputs (= 2 feature readings $\times$ 10 readings per second $\times$ 20 seconds). The dataset is split into three subsets for training (72.25\%), validation (12.75\%), and testing (15\%). The model consists of 4 hidden layers with ReLU activation functions (64, 32, 16, and 2 units). For training, the model uses the Sparse Categorical Cross Entropy loss function, the Adam optimizer (LR=0.00001), the checkpoint weight saving mechanism and early stopping with 64 epoch patience. The final model identifies an ongoing attack with $\approx$93.68\% classification accuracy.

Finally, model weights and biases are extracted, and the model is ported to ICSML using the methodology from Section \ref{sec:model_porting}.

\subsection{Attack Detection}
\begin{figure}[t]
    \centering
    \includegraphics[width=\columnwidth]{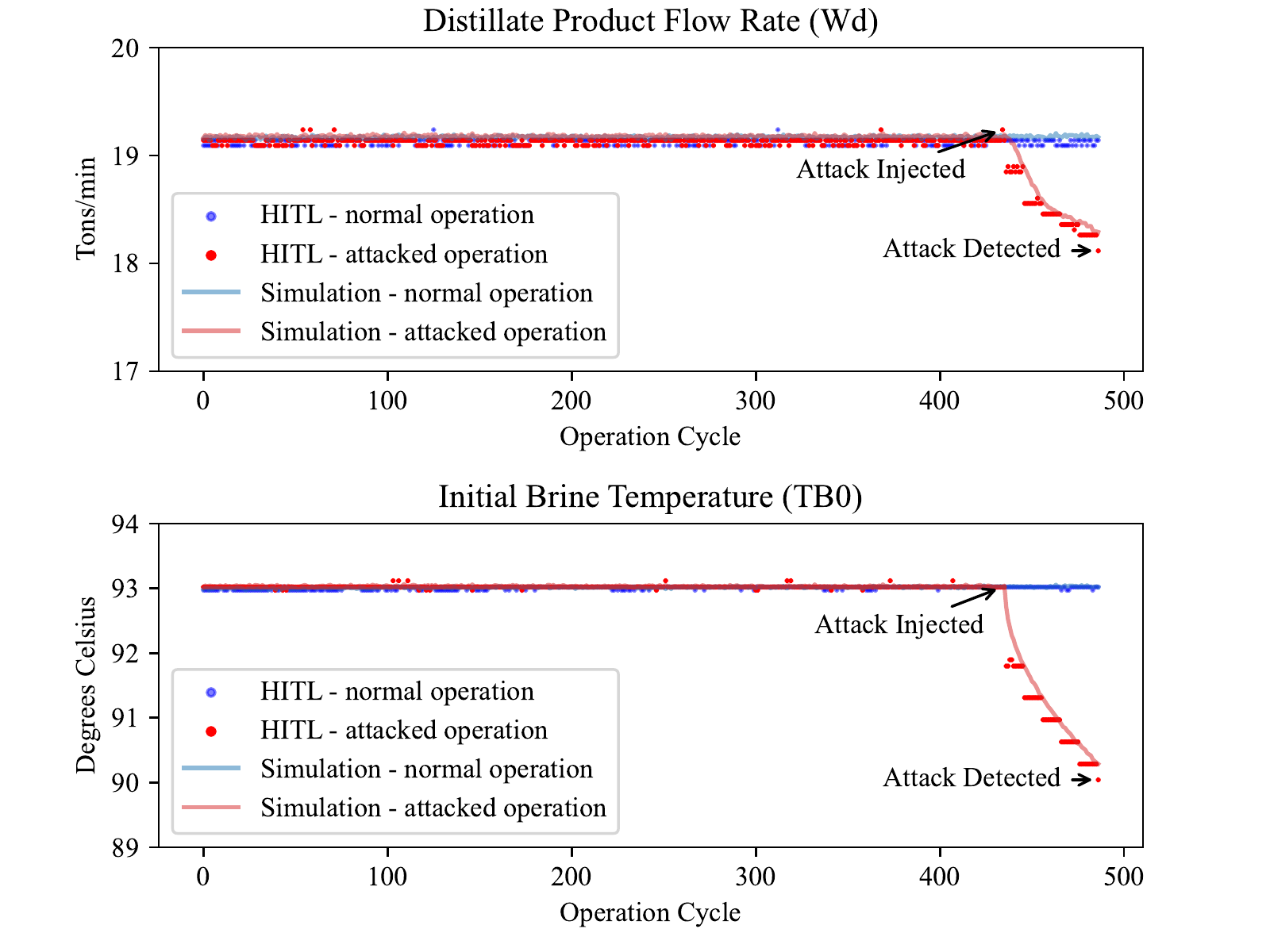}
    \caption{Distillate Product Flow Rate (Wd) and Initial Brine Temperature (TB0) timeseries collected from the PLC in the HITL MSF desalination plant. Blue points show sensor readings collected from the PLC under normal operation, and red from the attacked scenario time series. Sensor readings from the simulation are overlaid over the PLC data points. 
    }
    \label{fig:plc_defense_predictions}
\end{figure}

To test the effectiveness of the developed defense, we load the ICSML model onto the PLC, and, using the Simulink model, we simulate a series of process aware attacks that use parameters previously unseen by the model during its training. Figure \ref{fig:plc_defense_predictions} depicts Wd and TB0 sensor readings during normal operation and during a process aware attack scenario that sees the attacker tampering with the actuators regulating the recycle brine, steam, and water rejection flow rates, in order to negatively impact plant operation efficiency. Blue dots represent sensor readings as read by the PLC ADC during normal operation, while red dots are sensor readings recorded during the attack simulation. The light blue and red overlayed lines depict the sensor readings from the simulation during normal and attack operation scenarios respectively. The attack injection and detection points are annotated with arrows.

The attack is injected in the $436^{th}$ cycle of operation, and prom\-ptly detected 5 seconds later in the $486^{th}$ cycle. Detection does not happen instantaneously due to the classifier utilizing a sliding window model of operation which requires a certain number of malevolent sensor inputs to be triggered. This must also be kept under consideration when interpreting the model classification accuracy which is $\approx$93.68\% for any given set of inputs at a specific operation cycle and does not represent the accuracy for the detection of an attack overall. Through our experiments we also discovered that more subtle and gradual attacks can initially take longer to be recognized. This occurs because such attacks can often look like temporary stochastic benign anomalies in system operation. Nevertheless, improving the effectiveness of the defense method itself is outside the scope of this paper, as the focus here is on demonstrating the capabilities of ICSML. 

Attack mitigation tactics employed post detection differ based on the nature of the ICS system, and the type of the attack. For instance, attacks that target the PLC directly by tampering with its firmware, injecting false data or attempting denial of service, can be mitigated by having a secondary PLC run in parallel with the primary and automatically switching operations to it when an attack is detected \cite{proc:ml_sol1}. Also, ICS attacks involving actuator manipulation \cite{art:ukraine} can be mitigated using systems with complementary physical components.

Another interesting conclusion that can be drawn from Figure \ref{fig:plc_defense_predictions} is that the dataset for the ML model should be ideally collected on the PLC itself. The horizontal dot segments spotted in the graph, and the delta between the simulation produced sensor readings and PLC inputs, reveal some of the quantization effects, and loss in accuracy introduced by the internal PLC ADC. Since inference will happen on the PLC, training the defense model using inputs as processed by the PLC ADC will allow for more accurate model fine-tuning and generalization.

\subsection{Non-Intrusiveness} 
Non-intrusiveness is an essential property for an attack defense mechanism. To show that executing the ICSML defense model does not interfere with the cascading PID controller setup running on the PLC, inputs and outputs are recorded in controlled condition scenarios with and without the defense running on the PLC.

\begin{figure}[t]
    \centering
    \includegraphics[width=\columnwidth]{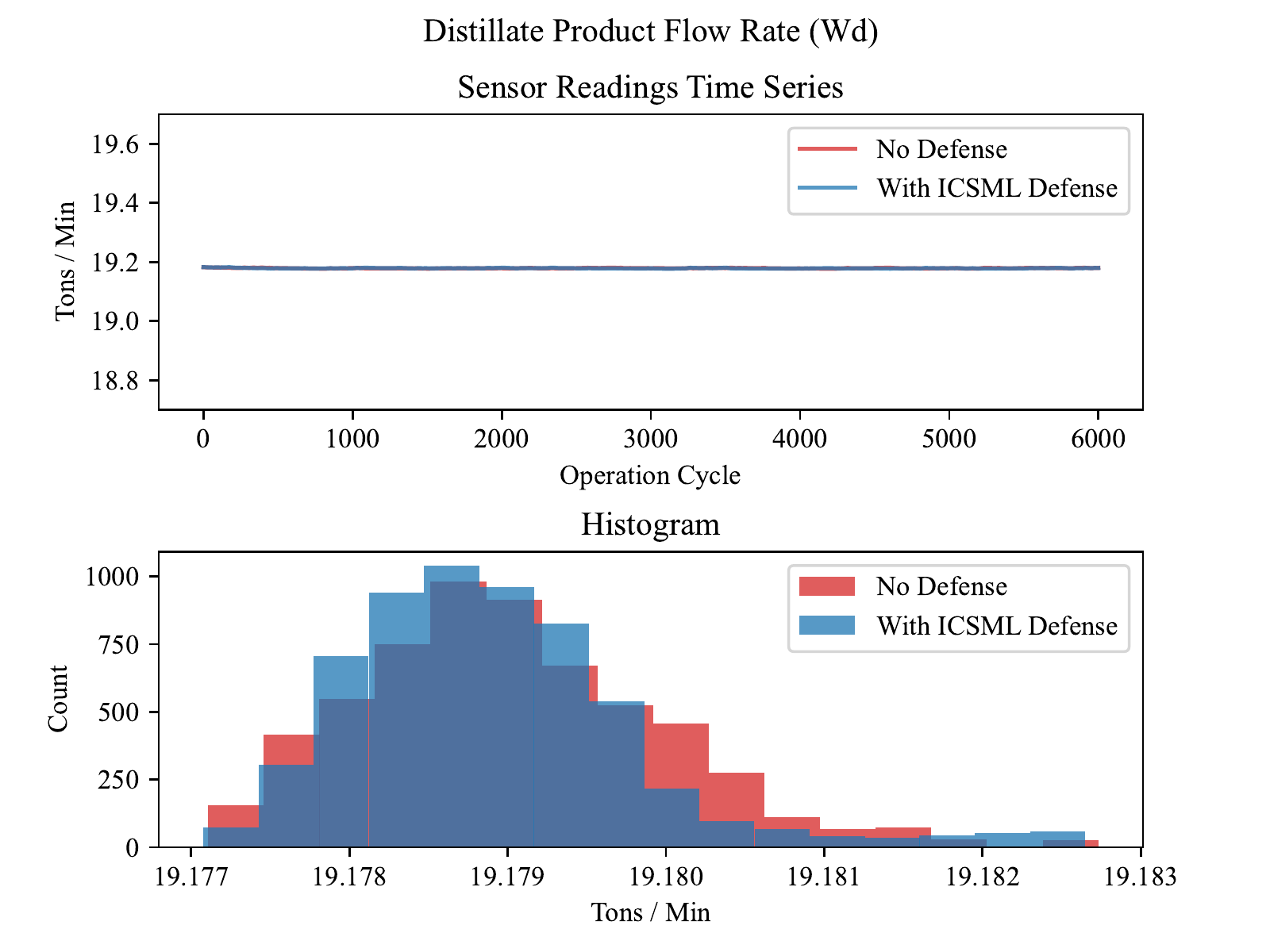}
    \caption{Time series and distribution of the distillate product flow rate as recorded in HITL setups where the ICSML defense was not and was present.}
    \label{fig:non_intrussive_wd}
    \vspace{-0.4cm}
\end{figure}

Figure \ref{fig:non_intrussive_wd} shows the Wd sensor inputs to the PLC for the first 6000 operation cycles after the desalination process has been initialized. In the scenario where the defense is not running, the recorded Wd time series has a mean of 19.18 tons/min, and a standard deviation of $9.47\cdot10^{-4}$. The Wd time series recorded while the defense is running on the PLC again has a mean of 19.18 tons/min, and standard deviation equal to $9.18\cdot10^{-4}$. Further inspection of the histograms in Figure \ref{fig:non_intrussive_wd} shows that Wd is similarly distributed in both cases. 

The variation in exact values of the process output can be attributed to the artificial additive noise, which is included in the simulation for added realism, and the noise introduced by the use of DAC and ADC converters in the HITL setup. Considering this, it can be understood that the execution of the ICSML model does not intrude upon the primary functionality of the PLC, and thus the defense does not affect the output of the process.

\section{Discussion \& Future Work}

\subsection{Performance}

Performance is an important aspect of ML inference on ICS as it defines the types of models that can run within the scan cycle time dictated by the controlled ICS process. ICSML can be further optimized for performance by exploring a number of avenues including automatically precompiling models to fully exploit weight pruning inference latency benefits. Additionally, non-linear functions in ICSML, like the Exponential Linear Unit activation function, can be rewritten in Taylor series format to improve performance. Furthermore, copy operations used during inference can be reimplemented on a case-by-case basis using vendor-specific low-level memory manipulation functions to ensure optimal performance. Finally, investigating the use of platform-specific libraries for optimized vector and matrix multiplications can potentially further decrease inference latency.

\subsection{Features}
In its current form, ICSML provides a complete set of components that allow the ML engineer to build and port densely connected Neural Networks to PLC software stacks. The existence of the concatenation layer allows building networks that branch out and merge, and also enables building Recurrent Neural Networks. Furthermore, ICSML includes the necessary components to build Convolutional Neural Networks (CNN).

Future expansion of ICSML can be inspired by the features found in established machine learning frameworks, such as Long-Short Term Memory (LSTM) and Gated Recurrent Units (GRUs) since PLCs often observe time series data as inputs. This would allow building more complex RNN models.

Beyond expanding the functionality of the framework itself, its usability can be extended through the development of external tools. As outlined in Section \ref{sec:model_porting}, ICSML code for ported ML models can be built in a structured process that follows well defined steps. This potentially allows the complete automation of the porting process by employing Model-To-Model Transformation techniques \cite{proc:model-to-model} that perform weight extraction and ICSML code generation using saved ML models exported directly from high-level ML frameworks. Additionally, this intermediate transformation tool can be further enhanced to offer and automatically apply suggestions for performance optimizations for a given ML model while taking into consideration the specific PLC hardware and the scan cycle parameters of the target ICS environment.

\subsection{Compatibility}
\label{sec:compatibility}
One of the most important objectives when developing the ICSML framework was cross-compatibility with PLCs from different vendors. This was achieved by leveraging the standardized programming environment and languages established by IEC 61131-3. ICSML code is built purely on IEC 61131-3 languages and, besides its binary data loading and storing functions, does not depend on vendor-specific code for its operation. Consequently, ICSML applications can be deployed on virtually any PLC compliant with the IEC 61131-3 standard. The framework has already been ported and tested on the Codesys V3 ecosystem, and the Beckhoff TwinCAT 3 platform. Porting the framework in the future to the software stacks of other vendors like SIEMENS STEP 7, WAGO e!Cockpit and Schneider Electric Control Expert, should be a trivial task.

\subsection{Security Applications}
ICSML enables a plethora of security applications beyond anomaly detection, which was discussed in the case study. ICSML can be used to implement intrusion detection systems that analyze the commands received by the PLC from other ICS devices. Additionally, it can enable ML methods that detect sensor reading spoofing and false data injection attacks, that attempt to manipulate the underlying process. PLCs are uniquely positioned to support such security applications because of their direct access to a variety of sensor readings. Also, using ICSML potentially reduces latency, and man-in-the-middle attack feasibility, since the data does not have to be sampled and transmitted over the network for inference purposes. Furthermore, inference running on the PLC can remove the need for external devices that sample and process data that the PLC has access to. As explored in Section \ref{sec:benchmarking}, the scan cycle, and capabilities of the available PLCs outline the complexity of the models that can be developed for such security applications.

\section{Related Work \& Applications}
The most directly comparable work to ICSML is Beckhoff's TwinCAT 3 proprietary platform \cite{beckhoff-ml}, which provides inference capabilities to the PLC by expanding the PLC framework stack with new components. Beckhoff's solution can potentially attain better performance than ICSML due to it having direct access to system resources, and it being able to utilize CPU and platform specific optimizations. However, as discussed in Section \ref{sec:compatibility}, ICSML natively runs on any IEC 61131-3 compatible device regardless of the manufacturer or underlying firmware, and does not need vendor support. Beckhoff's ML platform does not run on IEC 61131-3 languages; instead, it is regular software running in parallel on the PLC. Additionally, ICSML is open source software that does not require licensing, and can be freely customized and extended, whereas Beckhoff's ML platform is a closed ecosystem in which the user depends on the vendor for supporting new features, and requires purchasing a license to use. Therefore, with ICSML, the user only needs to update the control binary to add ML inference capabilities.

As discussed earlier in the paper, machine learning has been extensively explored for applications in ICS environments, especially for security purposes. 
In their work \cite{arc:encryption_plc}, Alves et al. use an open source PLC platform to embed an ML based intrusion prevention system onto the device, that thwarts network flood attacks like Denial of Service. 
Teixera et al. \cite{Teixeira} craft machine learning defenses for a water treatment and distribution testbed complete with a Supervisory Control and Data Acquisition (SCADA) system. Junejo et al. \cite{proc:Junejo} use unsupervised ML algorithms to create a behaviour-based defense for a water treatment facility testbed with PLCs. 
In their paper Yau et al. \cite{proc:yau} employ a semi-supervised one-class support vector machine ML algorithm to detect anomalous PLC events and aid forensics investigations. 
Meleshko et al. \cite{proc:ml_sol3} propose a method for detecting anomalous sensor data in cyber-physical systems and apply it on an example of a water supply system.

Apart from security solutions, ICSML can be used to implement other ancillary applications that are of interest in ICS environments, such as predictive maintenance. 
In their paper, Kanawaday and Sane \cite{kanawaday} utilize ML and IoT sensor data collected from an environment that includes PLCs, to predict possible failures and quality defects in a manufacturing process. 
Strauß et al. \cite{Straub} leverage ML and data from an industrial IoT environment with PLCs to enable predictive maintenance for an Electric Monorail System. 
Paolanti et al. \cite{Paolanti} propose an ML system for predictive maintenance using the Random Forest method and data collected from sensors and PLCs.

Beyond enabling auxiliary and passive functionality running side-by-side with primary tasks, ICSML can support ML applications created to directly control the underlying physical processes. As advancements in machine learning popularize its applications in a growing number of specialty areas, it is no surprise that ML techniques are actively considered in fields such as manufacturing \cite{ml-manufacturing-review}, process systems engineering \cite{wu2019machine} and energy systems \cite{sys-eng-ML-overview}. This growing trend of ML adoption in areas where ICS hardware is present foreshadows further use of ML in PLC control logic. 

\section{Conclusion}
In this work we studied the need for performing machine learning inference natively on PLCs. We explored the requirements and limitations for ICS machine learning applications running on PLCs and based on this we developed an ML framework using IEC 61131-3 languages. We then presented an end-to-end methodology for building and porting ML models to the developed framework. Afterwards, we explored the performance of the framework using a series of benchmarks, and compared it to the TensorFlow Lite inference framework. Finally, we showcased the abilities of the proposed framework and confirmed its non-intrusive nature by developing a real-life case study of an ML based defense for the MSF desalination process.

\section*{Resources}
ICSML is open source: \url{https://github.com/momalab/ICSML}

\bibliographystyle{ACM-Reference-Format}
\bibliography{icsml-bibliography}

\end{document}